\documentclass[a4paper]{article}

\usepackage{INTERSPEECH2020}

\usepackage{booktabs,multirow}
\usepackage{caption,subcaption}
\usepackage{tikz}
\usetikzlibrary{
    shapes.misc,
    positioning,
    chains,
    arrows,
    arrows.meta,
    decorations.pathmorphing,
    backgrounds,
    fadings,
    automata,
    calc,
}
\usepackage{tipa}
\usepackage{mathtools}

\DeclarePairedDelimiter{\abs}{\lvert}{\rvert}
\usepackage{url}

\title{Exploring TTS without T Using Biologically/Psychologically Motivated Neural Network Modules (ZeroSpeech 2020)}
\name{Takashi Morita, Hiroki Koda}
\address{
  Primate Research Institute, Kyoto University, JAPAN
  }
\email{tmorita@alum.mit.edu, koda.hiroki.7a@kyoto-u.ac.jp}

\begin{document}

\maketitle
\begin{abstract}
    In this study, we reported our exploration of Text-To-Speech without Text (TTS without T) in the Zero Resource Speech Challenge 2020, in which participants proposed an end-to-end, unsupervised system that learned speech recognition and TTS together.
    We addressed the challenge using biologically/psychologically motivated modules of Artificial Neural Networks (ANN), with a particular interest in unsupervised learning of human language as a biological/psychological problem.
    The system first processes Mel Frequency Cepstral Coefficient (MFCC) frames with an Echo-State Network (ESN), and simulates computations in cortical microcircuits.
    The outcome is discretized by our original Variational Autoencoder (VAE) that implements the Dirichlet-based Bayesian clustering widely accepted in computational linguistics and cognitive science.
    The discretized signal is then reverted into sound waveform via a neural-network implementation of the source-filter model for speech production.
\end{abstract}

\section{Introduction}

Recent developments in ANNs have drastically improved various tasks in natural language processing.
However, these developments are based on a large amount of annotated data, which are not available with respect to many minority languages and for real language-learning children.
Accordingly, unsupervised learning of languages based on raw speech data is required for industrial and academic purposes.
This study reports our exploration of TTS without T in the Zero Resource Speech Challenge 2020, which intended to develop an end-to-end, unsupervised system that can learn speech recognition and TTS together.
Even though the participants in the previous challenge mainly investigated mechanically sophisticated systems \cite{Dunbar_et_al_19_ZeroSpeech2019,Tjandra_et_al_19}, we addressed the challenge using biologically/psychologically motivated ANN modules, by considering the unsupervised learning of human language as a biological/psychological problem.
One of the adopted modules is our original discrete VAE that implements the Dirichlet-based Bayesian clustering within the end-to-end system.

\section{System Description}
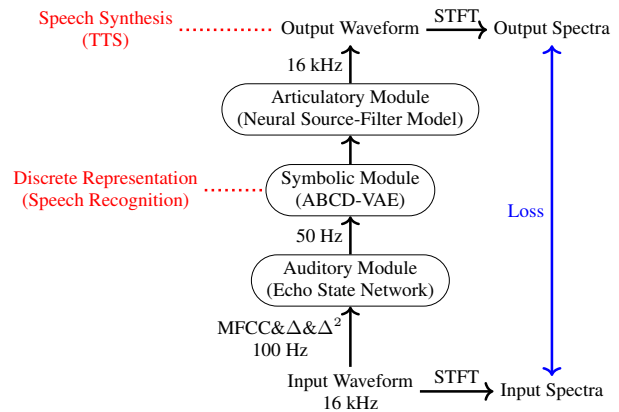
\begin{figure}
    \centering
    \scalebox{0.8}{
    \begin{tikzpicture}
        \node (input) [align=center] at (0,0) {Input Waveform\\16~kHz};
        \node (encoder) [draw, rounded rectangle, above=of input, align=center] {Auditory Module\\(Echo State Network)};
        \node (vae) [draw, rounded rectangle, above=2em of encoder, align=center] {Symbolic Module\\(ABCD-VAE)};
        \node (decoder) [draw, rounded rectangle, above=1.5em of vae, align=center] {Articulatory Module\\(Neural Source-Filter Model)};
        \node (output) [above=2em of decoder] {Output Waveform};
        \node (out_spectra) [right=of output] {Output Spectra};
        \node (in_spectra) at (input.center -| out_spectra.center) {Input Spectra};
        \draw [->, very thick] (input) to node [left,align=center] {MFCC\&$\Delta$\&$\Delta^2$\\100~Hz} (encoder);
        \draw [->, very thick] (encoder) to node [left] {50~Hz} (vae);
        \draw [->, very thick] (vae) to (decoder);
        \draw [->, very thick] (decoder) to node [left,align=center] {16~kHz} (output);
        \draw [->, very thick] (output) to node [above] {STFT} (out_spectra);
        \draw [->, very thick] (input) to node [above] {STFT} (in_spectra);
        \draw [<->, very thick, blue] (in_spectra) to node [left] {\color{blue}Loss} (out_spectra);
        \node (SR) [text=red,left=of vae,align=center] {Discrete Representation\\(Speech Recognition)};
        \node (TTS) [text=red,align=center] at (SR |- output) {Speech Synthesis\\(TTS)};
        \draw [-,very thick,dotted,red] (SR) to (vae);
        \draw [-,very thick,dotted,red] (TTS) to (output);
    \end{tikzpicture}
    }
    \caption{Overall architecture.}
    \label{fig:overall-architecture}
\end{figure}
Our network consisted of the auditory module (encoder), the symbolic module (discrete VAE), and the articulatory module (decoder; see Fig.~\ref{fig:overall-architecture}), each of which is described in details, below.
The dimensionality of the hidden layers of the network is 128, unless otherwise specified.

\subsection{Auditory Module: ESN} \label{sec:method_ESN}
The waveform data are first converted into 13 MFCCs (with 25~msec Hanning window and 10~msec stride), with their first and second derivatives concatenated.
Those frames are then processed by an ESN (with 2048 neurons) \cite{JaegerHaas04}.
ESNs are recurrent neural networks having sparse hidden-to-hidden connections (10\%), which are randomly initialized and fixed without updates via learning.
ESNs and their related model, liquid state machine, simulate computations in cortical microcircuits  \cite{Maass_et_al_02_LiquidStateMachine}.
We took every second output of the ESN and downsampled the signals from 100~Hz to 50~Hz.

\subsection{Symbolic Module: ABCD-VAE} \label{sec:method_ABCD-VAE}
\begin{figure}
    \centering
    \scalebox{0.8}{
        \begin{tikzpicture}
            \node (to_decoder) at (0,0) {To Articulatory Module};
            \node (jitter) [draw, rounded rectangle, below=1.5em of to_decoder] {Jitter};
            \node (weight_dot_value) [draw, circle, below=1.5em of jitter] {$\bullet$};
            \coordinate [right=of weight_dot_value] (value_to_weighted_sum);
            \coordinate [left=of weight_dot_value] (softmax_to_weighted_sum);
            \node (gumbel-softmax) [draw, rounded rectangle, align=center, below=2em of softmax_to_weighted_sum] {Gumbel-Softmax\\Sample};
            \node (softmax) [draw, rounded rectangle, below=2.2em of gumbel-softmax] {Softmax};
            \node (scale) [draw, rounded rectangle, below=1.5em of softmax] {Scale by $\sqrt{\updownarrow\mathbf{M}}$};
            \node (query_dot_key) [draw, circle, below=2em of scale] {$\bullet$};
            \coordinate (key_to_product) at (to_decoder.center |- query_dot_key.center);
            \coordinate [below=2em of key_to_product] (memory_to_key);
            \node (transpose) [draw, rounded rectangle] at (value_to_weighted_sum |- softmax.center) {Transpose};
            \coordinate (memory_to_value) at (value_to_weighted_sum |- memory_to_key);
            \coordinate (memory_x) at ($(memory_to_key)!0.5!(memory_to_value)$);
            \node (memory) [align=center, below=2em of memory_x] {Codebook\\$\mathbf{M}$};
            \coordinate (query_x) at ($2.0*(query_dot_key)-(key_to_product)$);
            \node (MLP) [draw, rounded rectangle] at (query_x |- memory_to_key) {MLP};
            \node (query) [align=center] at (query_x |- memory.center) {From Auditory Module\\$\mathbf{x}_i$};
            \draw [->, very thick] (memory) to (memory_x) to (memory_to_key) to node [left] {key} (key_to_product) to (query_dot_key);
            \draw [->, very thick] (memory_x) to (memory_to_value) to (transpose);
            \draw [->, very thick] (transpose) to node [right] {value} (value_to_weighted_sum) to (weight_dot_value);
            \draw [->, very thick] (query) to (MLP);
            \draw [->, very thick] (MLP) to node [left] {query} (query_x |- query_dot_key) to (query_dot_key);
            \draw [->, very thick] (query_dot_key) to node [left] {similarity} (scale);
            \draw [->, very thick] (scale) to (softmax);
            \draw [->, very thick] (softmax) to node (q_z) [left] {$q(z_i \mid \mathbf{x}_i)$} (gumbel-softmax);
            \draw [->, very thick] (gumbel-softmax) to node (z_tilde) [right] {$\mathbf{\tilde{z}}_i$} (softmax_to_weighted_sum) to (weight_dot_value);
            \draw [->, very thick] (weight_dot_value) to (jitter);
            \draw [->, very thick] (jitter) to (to_decoder);
            \coordinate [left=of q_z] (KL_to_q_z);
            \node (KL) [draw, rectangle, align=left] at (weight_dot_value.north -| KL_to_q_z) {Regularize by\\$D_{\textsc{kl}}[q(\boldsymbol{\pi}, \mathbf{z} \mid \mathbf{x}) \parallel p(\boldsymbol{\pi}, \mathbf{z})]$};
            \draw [->, dashed, very thick] (KL) to (KL_to_q_z) to (q_z);
        \end{tikzpicture}
    }
    \caption{ABCD-VAE, implemented as the scaled dot-product attention mechanism.
    The symbol ``$\updownarrow\mathbf{M}$'' represents the number of rows in the codebook matrix $\mathbf{M}$ (i.e., the dimensionality of the column vectors, set as 128) \cite{Vaswani_et_al_17_AttentionIsAllYouNeed}.}
    \label{fig:ABCD-VAE_architecture}
\end{figure}
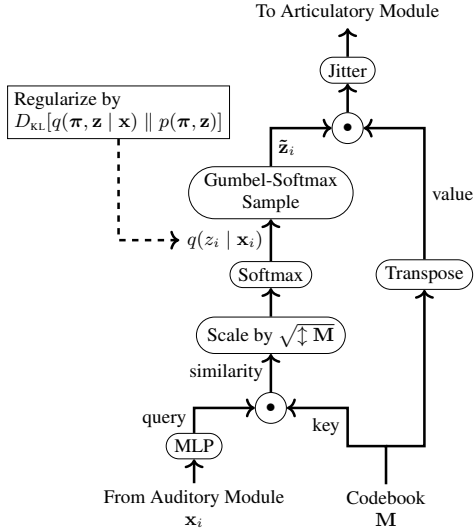
The output from the auditory module is a time series of real-valued vectors, and the role of the symbolic module is to discretize them.
Ideally, this module should classify sound frames into linguistically meaningful categories like phonemes whose utterance spans over the frames.
Since the size of phonemic inventory varies among languages, the symbolic module should also detect the appropriate number of categories instead of just putting frames into a predetermined number of classes.
A popular approach to such clustering problems in computational linguistics and cognitive science is Bayesian clustering based on the Dirichlet distribution/process \cite{Anderson90,KuriharaSato04,KuriharaSato06,Teh_et_al_06,KempPerforsTenenbaum07,Goldwater_et_al_09_word_seg,Feldman_et_al_13,Kamper_et_al_17,MoritaODonnell_sublexical_phonotactics}.
To build this Dirichlet-based clustering in our end-to-end system,
we propose a novel discrete VAE named the \emph{ABCD-VAE},
whose first four letters stand for the \underline{A}ttention-\underline{B}ased \underline{C}ategorical sampling with the \underline{D}irichlet prior
(simultaneously explored for analysis of birdsong in \cite{Morita_et_al_20_BF}).
The ABCD-VAE converts each output frame $\mathbf{x}_i$ from the auditory module to the probability $q(z_i \mid \mathbf{x}_i)$ of its classification to a discrete category $z_i$.
This mapping from $\mathbf{x}_i$ to $q(z_i \mid \mathbf{x}_i)$ is implemented by a Multi-Layer Perceptron (MLP) followed by a linear transform.

The prior on $z_i$ is the Dirichlet-Categorical distribution:
\begin{align}
    \boldsymbol{\pi} := (\pi_1, \dots, \pi_K)
        \sim&
            \mathrm{Dirichlet}(\boldsymbol{\alpha})
            \label{eq:dirichlet-generative}
            \\
    z_i \mid \boldsymbol{\pi}
        \sim&
            \mathrm{Categorical}(\boldsymbol{\pi})
            \label{eq:categorical-generative}
\end{align}
where the concentration $\boldsymbol{\alpha} := (\alpha_1, \dots, \alpha_K)$ is a free parameter and we set it as $\alpha_k=1$, $\forall k \in \{1,\dots,K\}$; $K = 256$ is an upper-bound for the number of frame categories to be used.

The classification probability $q(z_i \mid \mathbf{x}_i)$ is used to sample a one-hot-like vector $\mathbf{\tilde{z}}_i$ that approximates the categorical sample by the Gumbel-Softmax distribution \cite{Jang_et_al_17}.
The output of the ABCD-VAE is a continuous-valued vector given by $\mathbf{\tilde{z}}_i \mathbf{M}^{\textrm{T}}$, where $\mathbf{M}$ is a learnable matrix and $\mathbf{\tilde{z}}_i$ picks up one of its column vector, if it is indeed a one-hot vector.
One implementational trick here is that $\mathbf{M}$ is shared with the final linear transform yielding the classification probability $q(z_i \mid \mathbf{x}_i)$. Accordingly, the classification logits are given by the dot-product similarities between the transformation of $\mathbf{x}_i$ and the column vectors of $\mathbf{M}$.
This links the probability computation and the VAE's output just as the Vector-Quantized-VAE (VQ-VAE) does with the L2 distance \cite{vandenOord_et_al_17_VQVAE,Chorowski_et_al_19}, and makes the learning easier.
This shared-matrix architecture can be seen as the attention mechanism (having identical key and value)
\cite{Vaswani_et_al_17_AttentionIsAllYouNeed,Bahdanau_et_al_15_Attention,Devlin_et_al_18_BERT}
and, thus, we call it the attention-based categorical sampling (Fig.~\ref{fig:ABCD-VAE_architecture}).

We want to classify contiguous frames into the same category except when they include a phonemic boundary.
To encourage such classification, we adopted the jitter regularizer that randomly replaces each frame's category with one of its adjacent frames with probability, $0.12$ \cite{Chorowski_et_al_19}.

\subsection{Articulatory Module: Neural Source-Filter Model} \label{sec:method_source-filter}
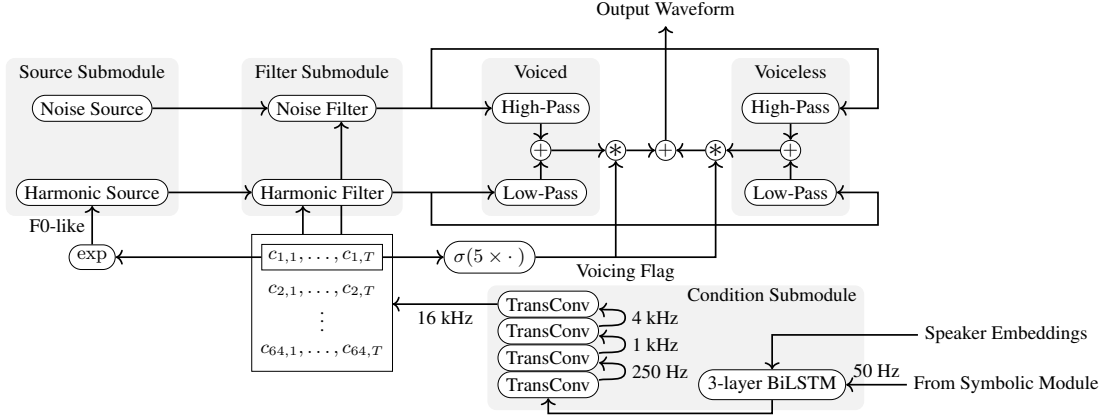
\begin{figure*}
    \centering
    \scalebox{0.8}{
    \begin{tikzpicture}[every node/.style={fill=white}]
    \node (source_noise) [draw, rounded rectangle] at (0,0) {Noise Source};
    \node (source_harmonic) [draw, rounded rectangle, below=3em of source_noise] {Harmonic Source};
    \node (source_title) [above=0.5em of source_noise, fill=none] {Source Submodule};
    \coordinate [yshift=-0.5em] (south) at (source_harmonic.south);
    \begin{scope}[on background layer]
		\filldraw[gray!10] ([xshift=-0.5em]source_harmonic.west |- source_title.north) [rounded corners] rectangle  ([xshift=0.5em]source_harmonic.east |- south);
	\end{scope}
    \node (filter_noise) [draw, rounded rectangle, right=6em of source_noise] {Noise Filter};
    \node (filter_harmonic) [draw, rounded rectangle] at (source_harmonic.center -| filter_noise.center) {Harmonic Filter};
    \draw [->, thick] (source_noise) to (filter_noise);
    \draw [->, thick] (source_harmonic) to (filter_harmonic);
    \node (filter_title) [fill=none] at (source_title.center -| filter_noise.center) {Filter Submodule};
    \begin{scope}[on background layer]
		\filldraw[gray!10] ([xshift=-0.5em]filter_harmonic.west |- filter_title.north) [rounded corners] rectangle  ([xshift=0.5em]filter_harmonic.east |- south);
	\end{scope}
    \node (HP_voiced) [draw, rounded rectangle, right=6em of filter_noise] {High-Pass};
    \node (LP_voiced) [draw, rounded rectangle] at (filter_harmonic.center -| HP_voiced.center) {Low-Pass};
    \node (HLplus_voiced) [draw, circle, inner sep=0pt] at ($(HP_voiced)!0.5!(LP_voiced)$) {$+$};
    \node (weight_voiced) [draw, circle, inner sep=0pt, right=2.8em of HLplus_voiced, scale=1.5] {$\ast$};
    \node (plus_total) [draw, circle, inner sep=0pt, right=1.5em of weight_voiced] {$+$};
    \node (weight_voiceless) [draw, circle, inner sep=0pt, scale=1.5] at ($2*(plus_total)-(weight_voiced)$) {$\ast$};
    \node (HLplus_voiceless) [draw, circle, inner sep=0pt] at ($(weight_voiceless)+(weight_voiced)-(HLplus_voiced)$) {$+$};
    \node (voiced_title) [fill=none] at (source_title.center -| HP_voiced.center) {Voiced};
    \begin{scope}[on background layer]
		\filldraw[gray!10] ([xshift=-0.5em]HP_voiced.west |- voiced_title.north) [rounded corners] rectangle  ([xshift=0.5em]HP_voiced.east |- south);
	\end{scope}
    \draw [->, thick] (filter_noise) to (HP_voiced);
    \draw [->, thick] (filter_harmonic) to (LP_voiced);
    \draw [->, thick] (HP_voiced) to (HLplus_voiced);
    \draw [->, thick] (LP_voiced) to (HLplus_voiced);
    \draw [->, thick] (HLplus_voiced) to (weight_voiced);
    \draw [->, thick] (weight_voiced) to (plus_total);
    \draw [->, thick] (HLplus_voiceless) to (weight_voiceless);
    \draw [->, thick] (weight_voiceless) to (plus_total);
    \node (HP_voiceless) [draw, rounded rectangle] at (HP_voiced.center -| HLplus_voiceless.center) {High-Pass};
    \node (LP_voiceless) [draw, rounded rectangle] at (filter_harmonic.center -| HP_voiceless.center) {Low-Pass};
    \node (voiceless_title) [fill=none] at (source_title.center -| HP_voiceless.center) {Voiceless};
    \begin{scope}[on background layer]
		\filldraw[gray!10] ([xshift=-0.5em]HP_voiceless.west |- voiceless_title.north) [rounded corners] rectangle  ([xshift=0.5em]HP_voiceless.east |- south);
	\end{scope}
    \coordinate (noise_to_voiceless1) at ($(filter_noise)!0.5!(HP_voiced)$);
    \coordinate (harmonic_to_voiceless1) at (noise_to_voiceless1 |- filter_harmonic.center);
    \coordinate [yshift=0.5em] (noise_to_voiceless2) at (noise_to_voiceless1 |- voiced_title.north);
    \coordinate (harmonic_to_voiceless2) at ([yshift=-0.5em]south -| harmonic_to_voiceless1);
    \coordinate [xshift=2em] (noise_to_voiceless3) at (HP_voiceless.east |- noise_to_voiceless2);
    \coordinate (harmonic_to_voiceless3) at (noise_to_voiceless3 |- harmonic_to_voiceless2);
    \coordinate (noise_to_voiceless4) at (HP_voiceless.east -| noise_to_voiceless3);
    \coordinate (harmonic_to_voiceless4) at (LP_voiceless.east -| harmonic_to_voiceless3);
    \draw [->, thick] (noise_to_voiceless1) to (noise_to_voiceless2) to (noise_to_voiceless3) to (noise_to_voiceless4) to (HP_voiceless);
    \draw [->, thick] (harmonic_to_voiceless1) to (harmonic_to_voiceless2) to (harmonic_to_voiceless3) to (harmonic_to_voiceless4) to (LP_voiceless);
    \draw [->, thick] (HP_voiceless) to (HLplus_voiceless);
    \draw [->, thick] (LP_voiceless) to (HLplus_voiceless);
    \node (output) [yshift=2em] at (plus_total |- noise_to_voiceless2) {Output Waveform};
    \draw [->, thick] (plus_total) to (output);
    \node (log_f0) [draw, rectangle, fill=none, below=2em of filter_harmonic] {$c_{1,1}, \dots, c_{1,T}$};
    \node (other_conds) [below=0em of log_f0, fill=none] {$
        \begin{array}{c}
         c_{2,1}, \dots, c_{2,T}\\
         \vdots \\
         c_{64,1}, \dots, c_{64,T}
        \end{array}
        $};
    \node (exp) [draw, rounded rectangle] at (log_f0.center -| source_harmonic.center) {$\exp$};
    \draw [->, thick] (log_f0) to (exp);
    \draw [->, thick] (exp) to node [left,fill=none] {F0-like} (source_harmonic);
    \coordinate [xshift=1em] (filter_noise_in) at (filter_noise.south);
    \coordinate [xshift=-1em] (filter_harmonic_in) at (filter_harmonic.south);
    \coordinate [yshift=0.5em] (condition_north) at (log_f0.north);
    \coordinate [xshift=0.5em] (condition_east) at (log_f0.east);
    \begin{scope}[on background layer]
		\draw ([xshift=-0.5em]log_f0.west |- condition_north) rectangle  ([yshift=-0.25em]condition_east |- other_conds.south);
		\draw [->, thick] (condition_north -| filter_noise_in) to (filter_noise.south -| filter_noise_in);
    	\draw [->, thick] (condition_north -| filter_harmonic_in) to (filter_harmonic.south -| filter_harmonic_in);
	\end{scope}
	\node (sigmoid) [draw, rounded rectangle, right=of log_f0] {$\sigma(5 \times \cdot\ )$};
	\draw [->, thick] (log_f0) to (sigmoid);
	\draw [->, thick] (sigmoid) to node [below,fill=none] {Voicing Flag} (sigmoid.center -| weight_voiceless.center) to (weight_voiceless);
	\draw [->, thick] (sigmoid.center -| weight_voiced.center) to (weight_voiced);
	\node (trans_conv_4) [draw, rounded rectangle, right=5em of other_conds, yshift=0.75em] {TransConv};
	\node (trans_conv_3) [draw, rounded rectangle, below=0em of trans_conv_4] {TransConv};
	\node (trans_conv_2) [draw, rounded rectangle] at ($2*(trans_conv_3)-(trans_conv_4)$) {TransConv};
	\node (trans_conv_1) [draw, rounded rectangle] at ($2*(trans_conv_2)-(trans_conv_3)$) {TransConv};
	\draw [->, thick] (trans_conv_4) to node [below,fill=none] {16~kHz} (trans_conv_4.center -| condition_east);
	\node (250Hz) [fill=none,right=4em of trans_conv_1.north] {250~Hz};
	\node (1kHz) [fill=none,anchor=west] at (trans_conv_2.north -| 250Hz.west) {1~kHz};
	\node (4kHz) [fill=none,anchor=west] at (trans_conv_3.north -| 250Hz.west) {4~kHz};
	\draw [->, thick] ([yshift=0.25em]trans_conv_1.east) to [out=10,in=270] (250Hz.west) to [out=90,in=350] ([yshift=-0.25em]trans_conv_2.east);
	\draw [->, thick] ([yshift=0.25em]trans_conv_2.east) to [out=10,in=270] (1kHz.west) to [out=90,in=350] ([yshift=-0.25em]trans_conv_3.east);
	\draw [->, thick] ([yshift=0.25em]trans_conv_3.east) to [out=10,in=270] (4kHz.west) to [out=90,in=350] ([yshift=-0.25em]trans_conv_4.east);
	\node (LSTM) [draw, rounded rectangle, right=of trans_conv_1, xshift=2em] {3-layer BiLSTM};
	\coordinate (global_south) at ([yshift=-0.8em]trans_conv_1.south);
	\draw [->, thick] (LSTM) to (global_south -| LSTM) to (global_south) to (trans_conv_1);
	\node (conditioner_title) [fill=none] at ([yshift=0.5em]LSTM.center |- trans_conv_4.center) {Condition Submodule};
	\begin{scope}[on background layer]
	    \filldraw[gray!10] ([xshift=-0.5em, yshift=0.25em]trans_conv_4.west |- trans_conv_4.north) [rounded corners] rectangle  ([xshift=0em, yshift=-0.5em]conditioner_title.east |- LSTM.south);
	\end{scope}
	\node (ABCD) [right=of LSTM] {From Symbolic Module};
	\node (speaker) [above=1em of ABCD] {Speaker Embeddings};
	\draw [->, thick] (ABCD) to node [above,fill=none] {50~Hz} (LSTM);
	\draw [->, thick] (speaker) to (speaker -| LSTM) to (LSTM);
    \end{tikzpicture}
    }
    \caption{The neural source-filter model, which customized the hn-NSF model in \cite{Wang_et_al_20_NeuralSourceFilter}.}
    \label{fig:source-filter}
\end{figure*}

The articulatory module receives the output from the symbolic module, and produces waveform from it.
A physiologically motivated model for the human voice production is the source-filter model.
We adopted a neural-network implementation of the source-filter model for the articulatory module (specifically, the hn-NSF in \cite{Wang_et_al_20_NeuralSourceFilter}; see Fig.~\ref{fig:source-filter}).

The articulatory module first processed the output from the ABCD-VAE with 3 layers of the bidirectional Long Short-Term Memory (LSTM), conditioned on two speaker embeddings; one of them is used as the initial hidden state and the other is concatenated with each frame of the output from the symbolic module.
The module then upsampled the outcome with four layers of transposed convolution (whose stride and kernel size were 5 and 25, respectively, for the bottom layer, and 4 and 16, respectively, elsewhere; the hidden dimensionality of the LSTM and transposed convolution was 128).
This yielded a 16~kHz sequence with 64~channels, which is termed $c_{j,t}$ in Fig.~\ref{fig:source-filter}.

The first channel of the upsampled sequence, $c_{1,1},\dots,c_{1,T}$, is intended to represent the base frequency of the output waveform in the log scale.
This F0-like channel is fed to the \emph{source} submodule that generated excitation signals from harmonic sine waves.
The \emph{filter} submodule transformed the resulting signals through 5 blocks of 10-layer dilated convolution (with 64 channels), conditioned on $c_{j,t}$.
The articulatory module had another source-filter flow designed for the production of noisy sounds, such as fricative consonants.
The noisy source is produced by a Gaussian and filtered through a single block of 10-layer dilated convolution.
(See \cite{Wang_et_al_20_NeuralSourceFilter} for the detailed architecture of the source and filter submodules.)

The harmonic and noisy outcomes are transformed by low-pass and high-pass Finite Impulse Response (FIR) filters, respectively, and combined additively.
We made two pairs of filters, one specialized for voiced sounds (0-5~kHz low-pass/stop and 7-8~kHz high-pass/stop bands) and the other for voiceless sounds (0-1~kHz low-pass/stop and 3-8~kHz high-pass/stop bands).
The filter coefficients are computed using the Remez exchange algorithm (with the order $10$) \cite{McClellanParks73,McClellan_et_al_73}.
The use of the voiced vs. voiceless sounds is switched by the F0-like signal;
$\textrm{F0}\gg 0$ flags the voiced sounds and $\textrm{F0}\approx 0$ flags the voiceless sounds.

\subsection{Training Objectives} \label{sec:method_objective}
The training objective, $\mathcal{L}$, of the network is given by, $\mathcal{L} := \bar{\mathcal{L}}_{\textsc{Spec}} + \mathcal{L}_{\textsc{kl}}$, which clearly consists of two terms.
The first term, $\bar{\mathcal{L}}_{\textsc{Spec}}$, compared the spectra of the input and output signals with three different time-frequency resolutions (Table~\ref{tab:freq-loss-resolution}).
The spectral loss between the input and output spectral sequences with $L$ frames and $M$ frequencies, $\mathcal{L}_{\textsc{Spec}}^{(M,L)}$ is defined by \cite{Wang_et_al_20_NeuralSourceFilter}:
\begin{align}
    \mathcal{L}_{\textsc{Spec}}^{(M,L)}
        :=&
            \frac{1}{2LM}
            \sum_{l=1}^{L}
            \sum_{m=1}^{M}
            \left(\log
                \frac{
                \abs{y_{m,l}}^2 + \epsilon
                }{
                \abs{\hat{y}_{m,l}}^2 + \epsilon
                }
            \right)^{2}
\end{align}
where $y_{m,l}$ and $\hat{y}_{m,l}$ are the input and output spectral sequences, respectively, and $\epsilon=10^{-5}$.
The average of $\mathcal{L}_{\textsc{Spec}}^{(M,L)}$ over the different time-frequency resolutions yielded the $\bar{\mathcal{L}}_{\textsc{Spec}}$.

\begin{table}
    \centering
    \caption{ Short-Time Fourier Transform (STFT) configurations of the spectral loss.}
    \label{tab:freq-loss-resolution}
    {\footnotesize
    \begin{tabular}{lrr@{\ }rr@{\ }r}
        \toprule
        FFT Config. Type&
            FFT Bins&
                \multicolumn{2}{c}{Frame Length}&
                    \multicolumn{2}{c}{Stride}\\
        \midrule
        Time-Dedicated&
            128&
                80&
                (5~ms)&
                    40&
                    (2.5~ms)\\
        Same as Input&
            512&
                400&
                (5~ms)&
                    100&
                    (5~ms)\\
        Freq.-Dedicated&
            2048&
                1920&
                (120~ms)&
                    640&
                    (40~ms)\\
        \bottomrule
    \end{tabular}
    }
\end{table}

The second term of the training objective, $\mathcal{L}_{\textsc{kl}}$, regularized the ABCD-VAE by the Kullback-Leibler (KL) divergence of the posterior probability, $q(\boldsymbol{\pi}, \mathbf{z})$ to the prior $p(\boldsymbol{\pi}, \mathbf{z})$ (cf. \cite{KingmaWelling14_VAE}).
To make this KL divergence computable, we adopted the mean-field variational inference and assumed that (i) $\boldsymbol{\pi}$ is independent of $\mathbf{z}$ and $\mathbf{x}$ in $q$ and (ii) $q(\boldsymbol{\pi})$ is a Dirichlet distribution \cite{Bishop06}.
Then, the KL divergence is rewritten as follows:
\begin{align}
    &\mathcal{D}_{\textsc{kl}}\left[
        q(\boldsymbol{\pi}, \mathbf{z})
        \parallel
        p(\boldsymbol{\pi}, \mathbf{z})
    \right]
        =
            \mathcal{D}_{\textsc{kl}}\left[
                q(\boldsymbol{\pi})
                \parallel
                p(\boldsymbol{\pi})
            \right]
            +
            \sum_{i=1}^{N} \mathcal{D}_{z_i}
            \\
    &\mathcal{D}_{\textsc{kl}}\left[
        q(\boldsymbol{\pi})
        \parallel
        p(\boldsymbol{\pi})
    \right]
        =
            \mathrm{E}_{q}\left[
                \log q(\boldsymbol{\pi})
            \right]
            -
            \mathrm{E}_{q}\left[
                \log p(\boldsymbol{\pi})
            \right]
            \\
    &\mathcal{D}_{z_i}
        :=
            \mathrm{E}_{q}\left[
                \log q(z_i \mid \mathbf{y}_i)
            \right]
            -
            \mathrm{E}_{q}\left[
                \log p(z_i \mid \boldsymbol{\pi})
            \right]
\end{align}
where all the expectations have a closed form.

The frame-based classification by the ABCD-VAE can result in an undesirable situation in which only categories spanning over many frames are detected, and short segments such as consonants are ignored due to the Occam's razor effect of the Dirichlet, prior.
To address this issue, we divided each $\mathcal{D}_{z_i}$ by $U_i$, the number of contiguous frames, including the one indexed by $i$, that yielded the same most probable category.
This weighting applied the Dirichlet-based clustering to spans of frames instead of individual frames, which ideally correspond to phonetic segments.
Accordingly, $\mathcal{L}_{\textsc{kl}}$ is defined for each data sequence as follows:
\begin{align}
    \mathcal{L}_{\textsc{kl}}
        :=
            \frac{1}{T}
            \left(
            \frac{S}{N}
            \mathcal{D}_{\textsc{kl}}\left[
                q(\boldsymbol{\pi})
                \parallel
                p(\boldsymbol{\pi})
            \right]
            +
            \sum_{i=1}^{S} \frac{1}{U_i} \mathcal{D}_{z_i}
            \right)
\end{align}
where $T$ is the number of samples in the waveform, $S$ is the number of frames in the discrete representation of the sequence, and $N$ is the total number of frames in the whole dataset.

The  concentration parameter, $\boldsymbol{\omega} := (\omega_1, \dots, \omega_K)$ of the Dirichlet, $q(\boldsymbol{\pi})$ is optimized by:
\begin{align}
    \omega_k
        =&
            \alpha_k + \sum_{i=1}^{N} q(z_i=k \mid \mathbf{x}_i)
            \label{eq:Dirichlet-optim_mean-field}
            \\
        =&
            \alpha_k + N \theta_k
            \label{eq:Dirichlet-optim_ABCD-VAE}
\end{align}
The second term on the right-hand side of Eq.~\ref{eq:Dirichlet-optim_mean-field} requires summation over all the $N$ frames, across different minibatches, and this is not efficient in minibatch learning.
Accordingly, we adopted Eq.~\ref{eq:Dirichlet-optim_ABCD-VAE}, instead  of Eq.~\ref{eq:Dirichlet-optim_mean-field}, where $\theta_k$ are learnable parameters of the ABCD-VAE such that $\sum_{k} \theta_k = 1$.

We trained the network for 36k iterations, using the Adam optimizer ($\beta_1=0.9$, $\beta_2=0.999$, $\epsilon=10^{-8}$) \cite{KingmaBa15_Adam}.
The learning rate is initially set as $4 \times 10^{-4}$ and halved at 16k, 24k, and 32k iterations.
During the first 4k iterations, we did not sample from the Gumbel-Softmax distribution in the ABCD-VAE, instead, we directly multiplied the classification probabilities, $q(\cdot \mid \mathbf{x}_i)$, and the transposed codebook, $\mathbf{M}^{\textrm{T}}$.
After this ``pretraining'' phrase, we annealed the temperature, $\tau$, of the Gumbel-Softmax following the equation, $\tau = \max\{0.5, \exp(-10^{-5} \iota)\}$, at an interval of 1k iterations, where $\iota$ counts the iterations \cite{Jang_et_al_17}.
Each batch consisted of 16 segments of speech sound whose duration was 1~s or shorter.
Those speech segments are randomly selected from the entire wav files when the files were longer than 1~s.
The network is implemented in PyTorch,%
\footnote{The code used in this study is available in \url{https://github.com/tkc-morita/ZeroSpeech2020_TTSwoT.git}.}
and ran on a private server with a single NVIDIA GeForce RTX 2080Ti graphic card.

\section{Results}
\begin{table*}
    \centering
    \caption{Scores for the speech recognition (encoding).}
    \label{tab:results_2019_StoT}
    {\footnotesize
    \begin{tabular}{llrrrrrrrr}
    \toprule
        \multirow{2}{*}{Language}&
            \multirow{2}{*}{Score Type}&
                \multirow{2}{*}{Baseline}&
                    \multirow{2}{*}{Topline}&
                        \multicolumn{2}{c}{Submitted Model}&
                            \multicolumn{2}{c}{CNN Encoder}&
                                \multicolumn{2}{c}{$+$F0 Learning}\\
        &
            &
                &
                    &
                        MAP&
                            Posterior&
                                MAP&
                                    Posterior&
                                        MAP&
                                            Posterior\\
    \midrule
        \multirow{2}{*}{English}&
            ABX&
                35.63&
                     29.85&
                        39.30&
                            35.46&
                                38.76&
                                    34.57&
                                        42.98&
                                            41.68\\
        &
            Bitrate&
                71.98&
                    37.73&
                        137.58&
                            405.37&
                                187.12&
                                    548.73&
                                        105.56&
                                            327.40\\
    \midrule
        \multirow{2}{*}{Surprise}&
            ABX&
                27.46&
                    16.09&
                        34.41&
                            18.87&
                                ---&
                                    ---&
                                        ---&
                                            ---\\
        &
            Bitrate&
                74.55&
                    35.20&
                        151.03&
                            ---&
                                ---&
                                    ---&
                                        ---&
                                            ---\\
    \bottomrule
    \end{tabular}
    }
\end{table*}
\begin{table}
    \centering
    \caption{Scores for the TTS (decoding).}
    \label{tab:results_2019_TtoS}
    {\footnotesize
    \begin{tabular}{llrrr}
    \toprule
        \multirow{2}{*}{Language}&
            \multirow{2}{*}{Score Type}&
                \multirow{2}{*}{Baseline}&
                    \multirow{2}{*}{Topline}&
                        Submitted\\
        &
            &
                &
                    &
                        Model\\
    \midrule
        \multirow{3}{*}{English}&
            MOS&
                2.14&
                    2.52&
                        1.19\\
        &
            CER&
                0.77&
                    0.43&
                        0.67\\
        &
            Similarity&
                2.98&
                    3.10&
                        1.14\\
    \midrule
        \multirow{3}{*}{Surprise}&
            MOS&
                2.23&
                    3.49&
                        1.77\\
        &
            CER&
                0.67&
                    0.33&
                        0.46\\
        &
            Similarity&
                3.26&
                    3.77&
                        1.22\\
    \bottomrule
    \end{tabular}
    }
\end{table}
The model performance for the speech recognition task (encoding) is evaluated on the ABX discriminability and bitrate of the embeddings.
The ABX discriminability scored the model's ability to distinguish phonetic minimal pairs;
The model is ran on minimal-paired signals, $A$ and $B$, and another signal from a different speaker, $X$, whose gold-standard transcription was identical to that of $A$. The ABX error rate is given by the probability that the model incorrectly assigned closer latent representation to $B$ and $X$ than to $A$ and $X$.
The bitrate evaluated the compression in the latent representation.

The ABX error rate of our proposed system was 39.30 for English and 34.41 for the surprise language \cite{Sakti_et_al_08_HMM,Sakti_et_al_08_A-STAR}, when the \emph{Maximum a Posteriori} (MAP) classification is scored using the Levenshtein distance and contiguous classmate frames are merged (Table~\ref{tab:results_2019_StoT}).
The error rates decreased to 35.46 and 18.87, respectively when the posterior probability (of the first frame of each classmate span) is scored using the KL divergence and the dynamic time warping (submitted as ``Auxiliary Embedding 1'').
These posterior-based scores are comparable to the baseline scores \cite{Dunbar_et_al_19_ZeroSpeech2019,Ondel_et_al_16,Wu_et_al_16_Merlin}, and even to the topline score of the surprise language \cite{Dunbar_et_al_19_ZeroSpeech2019}.
The gap between the MAP-based and posterior-based scores indicates that the MAP classification often ignored small acoustic differences in minimal pairs and their discrimination needed reference to non-MAP categories. For example, the model may assign the same MAP category to [\textsci] and [\textepsilon] but their second probable category may be different, say [i] vs. [e].

The simplest and, thus, possibly the weakest module of our model is the auditory module, which consisted of the ESN, or a random, non-trainable Recurrent Neural Network (RNN).
To evaluate this module, we replaced the ESN with a Convolutional Neural Network (CNN)-based encoder adopted in \cite{Chorowski_et_al_19} and tested it on the English data.
This replacement made only small improvements in the ABX scores, in compensation for the increased bitrates.

\begin{figure}
    \centering
    \subcaptionbox{
        Submitted Model
        \label{fig:F0-tracking_submitted}
    }{\includegraphics[width=0.98\columnwidth]{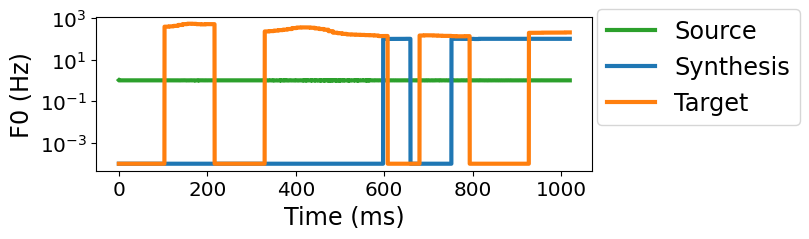}}
    \subcaptionbox{
        $+$F0 Learning
        \label{fig:F0-tracking_F0-learning}
    }{\includegraphics[width=0.98\columnwidth]{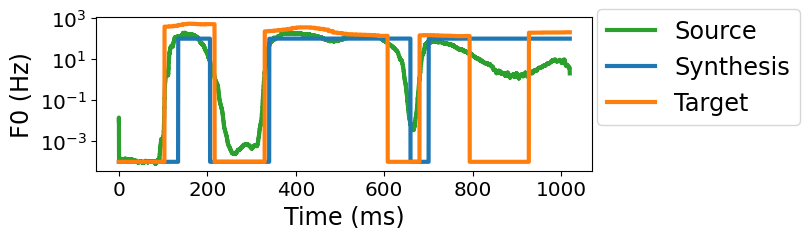}}
    \caption{Contours of F0 in the input of the source submodule (green), the speech synthesis (blue), and the target voice (orange). We used \texttt{V002\_4165536801.wav} in the voice dataset. F0 was zero when the speech sound was voiceless.}
    \label{fig:F0-tracking}
\end{figure}
The TTS performance is evaluated by three human-judged scores: Mean Opinion Score (MOS) on speech synthesis (greater values are better), Character Error Rate (CER) after human transcription of speech synthesis (the smaller, the better), and similarity to the target voice of speech synthesis (the greater, the better).
Our model got better CER scores than the baseline (Table~\ref{tab:results_2019_TtoS}), indicating that the encoder extracted important linguistic features and the decoder recovered them in the speech synthesis.
In contrast, the MOS and the similarity scores of our model were worse than the baseline.
We found that one major problem with the model was that it failed to learn the F0-like features used as the input to the source submodule (Fig.~\ref{fig:F0-tracking_submitted}), making the synthesized voice sound robotic.
This particular problem can be remedied by including the F0 loss in the objective function (Fig.~\ref{fig:F0-tracking_F0-learning}; the mean squared error between the F0 feature and ground truth is measured, and $\bar{\mathcal{L}}_{\textsc{Spec}}$ is replaced with the average of the F0 loss and it; F0 is computed using Perselmouth, a Python wrapper of Praat).
However, this F0 learning degraded the ABX scores (Table~\ref{tab:results_2019_StoT}), and the speech synthesis still sounded robotic to the authors.

\section{Discussion}
This study explored the TTS-without-T task using biologically/psychologically motivated modules of neural networks: the ESN for the auditory module, the ABCD-VAE for the symbolic module, and the neural source-filter model for the articulatory module.
Our technical contribution is the ABCD-VAE. It implemented the Dirichlet-based clustering in neural networks and made the end-to-end system possible.
Specifically, it automatically detects the statistically optimal number of frame categories (under an arbitrary upper bound) and enables more non-parametric learning than other discrete VAEs \cite{Jang_et_al_17,vandenOord_et_al_17_VQVAE,Chorowski_et_al_19}.

The ABCD-VAE yielded linguistically informative representation in the posteriorgrams, but the MAP representation missed some of this information.
This failure is likely not because of the limited capacity of the ESN encoder, which did not have any learnable parameters, since the canonical CNN-based encoder, adopted in previous work, yielded similar scores.
Similar problems were reported in last year's Zero Resource Speech Challenge;
most of the proposed discrete representations---particularly those with low bitrates---exhibited higher ABX error rates than the baseline, indicating a general difficulty in unsupervised learning of such representations \cite{Dunbar_et_al_19_ZeroSpeech2019}.

A bigger problem is found in the articulatory module---failure to learn the F0 feature used in the source submodule.
The naive learning of F0 improved this particular feature, but degraded the discrete representation of the ABCD-VAE and did not make the synthesized voice robustly more natural.
(Note that the original study of the neural source-filter model fed the gold-standard F0 to the model and thus the F0 learning was not an issue.)
Thus, successful training of this articulatory module---in the end-to-end setting---will be the central issue for the next Zero Resource Speech Challenge in our framework.

\section{Acknowledgements}
We gratefully acknowledge the support of
MEXT Grant-in-aid for Scientific Research on Innovative Areas \#4903 (Evolinguistic; JP17H06380) and the JST Core Research for Evolutional Science and Technology 17941861 (\#JPMJCR17A4).




\end{document}